\documentclass{article}

\PassOptionsToPackage{numbers, compress}{natbib}
\usepackage[preprint]{nips_2018}

\usepackage[utf8]{inputenc} % allow utf-8 input
\usepackage[T1]{fontenc}    % use 8-bit T1 fonts
\usepackage{hyperref}       % hyperlinks
\usepackage{url}            % simple URL typesetting
\usepackage{booktabs}       % professional-quality tables
\usepackage{amsfonts}       % blackboard math symbols
\usepackage{nicefrac}       % compact symbols for 1/2, etc.
\usepackage{microtype}      % microtypography
\usepackage{booktabs} % for professional tables
\usepackage{hyperref}
\usepackage{pgfplotstable}
\usepgfplotslibrary{groupplots,colorbrewer}
\pgfplotsset{compat=1.13}
\pgfplotstableset{col sep=comma}

\usepackage{amssymb}
\usepackage{amsmath}

\usepackage{natbib}

\usepackage{graphicx} % more modern
\usepackage{wrapfig}
\usepackage{subfigure} 
\usepackage{multirow}
\usepackage{adjustbox}

\usepackage{listings}
\usepackage{textcomp}

\usepackage{tikz}
\usepackage[most]{tcolorbox}

\usepackage[framemethod=TikZ]{mdframed}
\usepackage{xcolor}
\newcounter{example}
\renewcommand{\theexample}{\thesection.\arabic{example}}
\mdfdefinestyle{example}{%
    linecolor=blue,
    outerlinewidth=2pt,
    bottomline=false,
    leftline=false,rightline=false,
    skipabove=\baselineskip,
    skipbelow=\baselineskip,
    frametitle=\mbox{},
}
\newmdenv[%
    style=example,
    settings={\global\refstepcounter{example}},
    frametitlefont={\bfseries Example~\theexample\quad},
]{example}
\newmdenv[%
    style=example,
    frametitlefont={\bfseries Example~\quad},
]{example*}

\usepackage{algorithm}
\usepackage{algorithmic}

\makeatletter
\makeatletter
\newcommand*{\da@rightarrow}{\mathchar"0\hexnumber@\symAMSa 4B }
\newcommand*{\da@leftarrow}{\mathchar"0\hexnumber@\symAMSa 4C }
\newcommand*{\xdashrightarrow}[2][]{%
  \mathrel{%
    \mathpalette{\da@xarrow{#1}{#2}{}\da@rightarrow{\,}{}}{}%
  }%
}
\newcommand{\xdashleftarrow}[2][]{%
  \mathrel{%
    \mathpalette{\da@xarrow{#1}{#2}\da@leftarrow{}{}{\,}}{}%
  }%
}
\newcommand*{\da@xarrow}[7]{%
  \sbox0{$\ifx#7\scriptstyle\scriptscriptstyle\else\scriptstyle\fi#5#1#6\m@th$}%
  \sbox2{$\ifx#7\scriptstyle\scriptscriptstyle\else\scriptstyle\fi#5#2#6\m@th$}%
  \sbox4{$#7\dabar@\m@th$}%
  \dimen@=\wd0 %
  \ifdim\wd2 >\dimen@
    \dimen@=\wd2 %   
  \fi
  \count@=2 %
  \def\da@bars{\dabar@\dabar@}%
  \@whiledim\count@\wd4<\dimen@\do{%
    \advance\count@\@ne
    \expandafter\def\expandafter\da@bars\expandafter{%
      \da@bars
      \dabar@ 
    }%
  }%  
  \mathrel{#3}%
  \mathrel{%   
    \mathop{\da@bars}\limits
    \ifx\\#1\\%
    \else
      _{\copy0}%
    \fi
    \ifx\\#2\\%
    \else
      ^{\copy2}%
    \fi
  }%   
  \mathrel{#4}%
}
\makeatother

\newcommand{\cind}[0]{\mathrel{\text{\scalebox{1.07}{$\perp\mkern-10mu\perp$}}}}
\newcommand{\ncind}[0]{\mathrel{\text{\scalebox{1.07}{$\not\perp\mkern-10mu\perp$}}}}

\title{Causal Interventions for Fairness}
\author{
  Matt J. Kusner \\
  The Alan Turing Institute \\
  University of Warwick \\
  \texttt{mkusner@turing.ac.uk}
  \And
  Chris Russell \\
  University of Surrey \\
  The Alan Turing Institute \\
  \texttt{crussell@turing.ac.uk}
  \AND
  Joshua R. Loftus \\
  New York University \\
  \texttt{loftus@nyu.edu}
  \And
  Ricardo Silva \\
  University College London \\
  The Alan Turing Institute \\
  \texttt{ricardo@stats.ucl.ac.uk}
}

\begin{document}

\maketitle
   
\begin{abstract}
%A substantial limitation of algorithmic fairness, i.e. approaches designed to guarantee that decisions made by algorithms are fair and equitable, is that these interventions guaranteeing fairness are too little and too late. By the time a decision is made, many protected subgroups have already suffered from discrimination and lost opportunities for a substantial period of time. Rather than relying on private companies and other decision-making entities  to re-balance past mistakes, it would be more effective for charities and government organizations to make early interventions to stop opportunities from being lost in the first place. We provide a novel formulation for allocating such interventions, to choose them in a way to maximize their positive effects while improving the fairness of the overall system.  
%Hey Matt, I remembered I already had preliminary text for an abstract, so pasted it in.
%Feel free to change
%It's really nice!
%Josh changed to below, feel free to revert/combine/modify
Most approaches in algorithmic fairness constrain machine learning methods so the resulting predictions 
%or decisions 
satisfy %a definition of fairness.
%formalization 
one of several intuitive notions of fairness. 
While this may help private companies comply with non-discrimination laws or avoid negative publicity, we believe it is often too little, too late. By the time the training data is collected, individuals in disadvantaged groups have already suffered from discrimination and lost opportunities due to factors out of their control. In the present work we focus instead on \emph{interventions} such as a new public policy, and in particular, how to maximize their positive effects while improving the fairness of the overall system. We use causal methods to model the effects of interventions, allowing for potential \emph{interference}--each individual's outcome may depend on who else receives the intervention. We demonstrate this with an example of allocating a budget of teaching resources using a dataset of schools in New York City.
\end{abstract}

% ======== %
% OUTLINE: %
% ======== %
% I. Introduction
%    a) Fairness
%    b) Prediction vs. Intervention
% II. Background
%	 a) Causality
%    b) Counterfactual Fairness
%    c) Interference in Causality
% III. Setup
%	 a) Objective
%	 b) Constraints
% IV. Experiments
%	 a) Describe data
%	 b) Linear model
%	 c) Max model
% V. Related Work
%	 a) Fairness and interventions
%		- FATML paper, Elias Paper, Delayed Impact paper
%	 b) Interference work
% VI. Conclusion

\section{Introduction}
% (machine learning is reinventing the world)
% ML is hugely successful
% so people use it everywhere
% including life-changing decisions
% ML can give people an opportunity or deny them it
% OR
% Machine learning has entered a new era. It
Machine learning is used by companies, governments, and institutions to make life-changing decisions about individuals, such as how much to charge for insurance \cite{peters2017statistical}, how to target job ads \cite{yang2017combining}, and who may likely commit a crime \cite{zeng2017interpretable}. Thus, it can be used to give people an opportunity or deny it. 

% prediction
%Until recently, there had been few investigations into the predictions these models gave to people of different races, genders, sexual orientations, or otherwise. However,
Recently, a number of
striking examples of bias against sensitive attributes such as race or gender have raised awareness about potential downsides of algorithmic decisions. For example, Google's advertisement system was more likely to produce ads implying a person had been arrested when the search term was a name commonly associated with African Americans \cite{sweeney2013discrimination}. In another case, algorithms that learn word embeddings from news articles resulted in sexist associations such as ``woman'' being associated with ``homemaker'' \cite{bolukbasi2016man}.

% related work
Partially in response to these examples, there has been much recent work aimed at quantifying and removing these biases %\cite{kamiran2009classifying,kamishima2012fairness,dwork2012fairness,zemel2013learning,edwards2015censoring,bolukbasi2016man,flores2016false,kleinberg2016inherent,hardt2016equality,chouldechova2017fair,pleiss2017fairness,Berk2017,zafar2017fairness2,kusner:17,kilbertus2017avoiding,nabi2017fair,zhang2018fairness,dwork2018decoupled,liu2018delayed}.
\cite{Berk2017,bolukbasi2016man,chouldechova2017fair,dwork2012fairness,dwork2018decoupled,edwards2015censoring,hardt2016equality,kamiran2009classifying,kamishima2012fairness,kilbertus2017avoiding,kleinberg2016inherent,kusner:17,larson2016we,liu2018delayed,nabi2017fair,pleiss2017fairness,zafar2017fairness2,zemel2013learning,zhang2018fairness}. In large part, these works have used observational data to quantify fairness with various measures and correct for it. Unfortunately, it is rarely clear which measure is right for a given problem, and we cannot resort to combining them as many are mutually incompatible \cite{kleinberg2016inherent,pleiss2017fairness}. Alternatively, causal approaches \cite{kilbertus2017avoiding,kusner:17,nabi2017fair,zhang2018fairness} allow the modeler to design customized fairness measures via domain-specific causal graphs.

%impact
% if we really believe that ML is going to change the world then we have an opportunity to make the world fairer, we cannot pass this up 
%Previously, \cite{pmlr-v81-barabas18a} have called for an approach based on intervention rather than prediction. The present work answers this call, drawing upon the well-established ...
However, in this work we argue, that if machine learning is now really changing people's lives with its decisions, then we have an opportunity to not only make fair predictions, \emph{but to alter the unfairness in the system itself}. % make the world fairer.
In agreement with \cite{pmlr-v81-barabas18a} we argue in favor of exploiting interventions to mitigate associations between important outcomes and sensitive attributes (i.e., race, sex, gender identity, sexual orientation, or otherwise). 
To do so we draw upon the well-established tools of causal inference for interventions. Inspired by the causal notion of \emph{counterfactual fairness} \cite{kusner:17} we design counterfactual quantities to measure how much an intervention weakens the association between an outcome and a sensitive attribute. These quantities measure how much an individual benefits from an intervention \emph{purely because} they are members of a privileged group. We introduce an optimization procedure to find an intervention maximizing total benefit that simultaneously weakens this privileged association. 
%without allocating resources to individuals who would benefit more \emph{because} they are members of a privileged group. 
We demonstrate our method on a real-world dataset to assess the impact of funding advanced classes on college-entrance exam-taking.

\section{Background: Counterfactual Fairness}

\begin{figure*}[t]
\centering
\includegraphics[width=\textwidth]{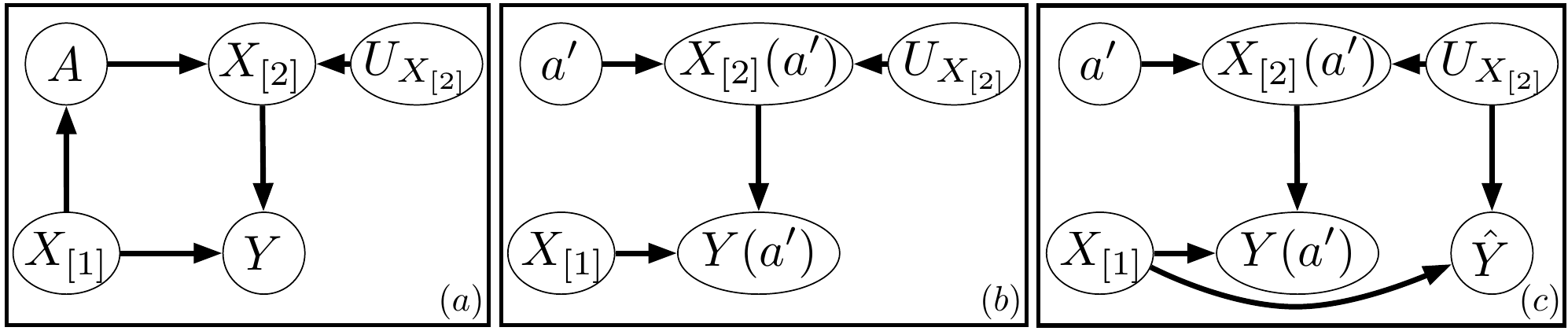}
\vspace{-3ex}
\caption{(a) A simple causal graph with two features $X_{[1]}$, $X_{[2]}$ besides the protected attribute $A$ and outcome $Y$. Variables named $U$ represent hidden variables. (b) A counterfactual system representing the fixing $A$ to some value $a'$, explicitly showing new vertices where necessary: vertices ``$V(a')$'' are labeled ``$V$'' whenever they are not descendants of $A$. (c) The same graph, augmented by a choice of $\hat Y$ that does not change across counterfactual levels.}
\label{fig:cf_basic}
\end{figure*}

Counterfactual fairness \cite{kusner:17} is a definition of fair predictors that is based in causal models. Let $A$ be a (set of) protected attribute(s), $Y$ an outcome of interest and $X$ a set of other features. A predictor $\hat Y$ of $Y$ satisfies counterfactual fairness if it satisfies the following criteria:
\begin{equation}
P(\hat Y(a) = y\ |\ A = a, X = x) = P(\hat Y(a') = y\ |\ A = a, X = x),
\label{eq:fairness}
\end{equation}
\noindent for all $a, a', y, x$ in the domains of $A$, $Y$, and $X$. The notation $V(a')$ represents a \emph{counterfactual} variable corresponding to a \emph{factual} variable $V$.\footnote{Our notation is slightly different but equivalent to the one in \cite{kusner:17}.} It represents the counterfactual statement ``the value of $V$ had $A = a'$ instead of the factual value''. As used by \cite{kusner:17}, counterfactuals are defined within Pearl's Structural Causal Model (SCM) framework \cite{pearl:00}. This framework defines a causal model by a set of \emph{structural equations} $V_{[i]} = g_{[i]}(pa_{[i]}, U_{[i]})$, which correspond to a directed acyclic graph (DAG) $\mathcal G$ where $pa_{[i]}$ are the observable parents of $V_{[i]}$ in $\mathcal G$, and $U_{[i]}$ is (set of) parent-less unobserved latent causes of $V_{[i]}$.\footnote{We use boxed subscripts $[i]$ to index the $i$th variable/feature and unboxed subscripts $j$ to denote the $j$th individual. Also, an edge from a set of variables $X$ to another variable $Y$ will mean each feature in $X$ causes $Y$.} The counterfactual ``world'' is generated by fixing $A$ to $a'$, removing any edges into vertex $A$, and propagating the change to all descendants of $A$ in the DAG, as shown in Figure~\ref{fig:cf_basic} (a), (b). Any variables in the model that are not in $A \cup X$, and are not descendants of $A$, can be inferred given the event $\{A = a, X = x\}$, as the remaining set of equations defines a joint distribution. 

The motivation behind (\ref{eq:fairness}) is that the protected attribute should not be a cause of the predicted outcome for any particular individual, other things being equal (in this case, the non-descendants of $A$ in the DAG). Informally, it translates as ``we would not make a different prediction for this person had this person's protected attribute been different, given what we know about them'' (the prediction is probabilistic if it depends on unobserved variables). This is in contrast to non-causal definitions which enforce observational criteria such as $Y \cind A \ |\ \hat Y$ (calibration \cite{flores2016false}), or $\hat Y \cind A\ |\ Y$ (equalized odds \cite{hardt2016equality}). As discussed by \cite{chouldechova2017fair,kleinberg2016inherent}, in general it is not possible to enforce both conditions, particularly if $A \ncind Y$. This will happen if $A$ is a cause of $Y$: in a SCM if $A$ is an ancestor of $Y$ in the DAG. In a nutshell, counterfactual fairness can be interpreted as building a predictor $\hat Y$ that is not a descendant of $A$ if we augment the system with a vertex representing the predictor, as in Figure \ref{fig:cf_basic} (c). Within the family of predictors satisfying such dependencies, predictive accuracy of $\hat Y$ with respect to $Y$ is maximized. The recent survey \cite{loftus:18} provides an extensive overview of causal thinking in fairness problems and the role of counterfactual fairness in particular.

% \begin{figure}[t]
% \begin{center}
% \emph{TODO (Figure (a) should be a graph $X_1 \rightarrow Y, X_1 \rightarrow A, A \rightarrow X_2, X_2 \rightarrow Y$, and the respective, $U_{X_2}$. Figure (b) should be a graph like (a), but with nodes $A$, $X_2$ and $Y$ renamed as $a'$, $X_2(a'), Y(a')$ and the edge $X_1 \rightarrow Y$ removed. (c) should be a graph like (b), but with the extra node $\hat Y$ and edges $X_1 \rightarrow \hat Y$, $U_{X_2} \rightarrow \hat Y$)}
% \end{center}
% \end{figure}

In this formulation, the world itself might be unfair in the sense that $Y$ is caused by $A$, but we have the freedom of setting our predictor in a way where $\hat Y$ is not causally affected by the protected attribute. In an important sense, this only addresses an injustice in the world in an indirect way. If $A$ is ``race'', $Y$ is ``person will default a loan'', and $A$ is a cause of $Y$, counterfactual fairness emphasizes making predictions of loan default that excludes information about unfair events. %(for instance, in case society is such that being of a particular race decreases the chance of finding a job), 
But it can only do so indirectly, and hope that this makes a change in the long run in the causal paths from $A$ to $Y$. Formalizing the interplay between $\hat Y$ and the changes in the causal graph would require longitudinal data on the adoption of $\hat Y$ and/or many assumptions about the dynamics of society. While such issues are just starting to be considered in the literature (we are aware of only one other work \cite{liu:18}, which is complimentary to ours), our goal here is to directly distinguish the relationship between $A$ and $Y$ to have immediate impact. To do so, our approach is to leverage existing work in causal interventions, and to link them to formal counterfactual definitions of fairness. We introduce our approach below. %, which we introduce here.
% we would like to leverage existing interventions that bypass defining new predictors, have immediate impact, and which can be estimated using existing techniques. What is missing is how to link them to formal counterfactual definitions of fairness, which we introduce here.

\section{Problem Formulation and Solution}

We consider the \emph{interventional problem}, which complements the prediction problem described in the previous section. In this scenario, we assume that we have the opportunity of altering the existing relationship between $A$ and $Y$ by performing interventions in the system. Within the SCM framework, the concept of ``perfect intervention'' is one of its primitives: modifications in the causal process of the world that can be represented by breaking edges in the causal graph. For example, if it was possible to perform a perfect intervention on $X_{[2]}$ in the graph of Figure \ref{fig:cf_basic}(a), this would imply the deletion of the edge $A \rightarrow X_{[2]}$. No propagation from $a'$ to $Y$ would occur in the graph of Figure \ref{fig:cf_basic}(b). 

Perfect interventions are often impossible in real problems in social science. Otherwise, a direct intervention on $Y$ would solve the problem. Instead, we consider ``soft,'' or imperfect interventions that alter the relationship between $A$ and $Y$ without removing the pathways. As commonly done in the literature \cite{sgs:93,pearl:00,dawid:02}, we can represent interventions as special types of vertices in a causal graph which index particular counterfactuals. For instance, if each individual $i$ is given a particular intervention $Z_i \!=\! z_i$, we can represent its counterfactual outcomes as $Y_i(z_i)$, and the corresponding causal graph will include a vertex $Z$ pointing to $Y$. This vertex does not represent a random variable, but the index of a choice of intervention. In particular, we will adopt the convention ``$Z_i \!=\! 0$'' to denote that no intervention is applied to unit $i$: instead, $Z_i = 0$ denotes that we let the $i$th instantiation of the system run its ``natural regime.'' In Figure \ref{fig:cf_basic}(a), we have a graph defined to represent the natural regime of a system. We could define $Z = $``set variable $X_{[2]}$ to $x_{[2]}$'' as the perfect intervention on $X_{[2]}$, adding an edge from $Z$ to $X_{[2]}$. In general we could have $Z$ being a parent to all vertices, with $Z = z$ representing a particular choice of conditional distribution for each vertex given their parents.

\subsection{Assumptions}

We consider \emph{interference} models, where interventions applied to one individual affect other individuals \cite{sobel:06,betsy:14}. That is, it is allowed that $Z_i \ncind Y_j$ for $i \neq j$. As in \cite{aronow:17}, we will not be concerned about direct causal connections between different outcomes $\{Y_i, Y_j\}$, focusing exclusively on the intention-to-treat effects of $\{Z_1, Z_2, \dots, Z_n\}$ on $\{Y_1, Y_2, \dots, Y_n\}$, where $n$ is the number of individuals.

As discussed in the previous section, each individual $i$ has a set of features represented as a vector $X_i$, and each individual belongs to a sensitive group $A_i$. For each individual $i$, we decide to perform intervention $Z_i$. For simplicity of presentation, we will assume throughout that each $A_i, Z_i$ are binary, with $Z_i \!=\! 0$ representing the ``idle'' choice of making no direct intervention on $i$. In contrast to the usual definition of counterfactual fairness where the only counterfactual index is given by the protected attributes, we will use $Y_i(a_i, \mathbf z)$ to denote the counterfactual outcome for individual $i$ with a fixed protected attribute $A_i = a_i$ and control signal $\mathbf z \equiv [z_1, z_2, \dots, z_n]^\intercal$ where $z_i$ is the assignment to intervention variable $Z_i$. In particular $\{Y_1(a_1, \mathbf 0), \dots, Y_n(a_n, \mathbf 0)\}$ is the set of outcomes where we decide to leave the system unperturbed.

In a causal graph, each outcome $Y_i$ is assumed to be directly influenced by the individual's sensitive group $A_i$, features $X_i$, and the intervention $Z_i$, represented by the edges:
\begin{align}
A_i \rightarrow Y_i \;\;\;\;\;\; X_i \rightarrow Y_i \;\;\;\;\;\; Z_i \rightarrow Y_i. \nonumber
\end{align}
These correspond to inputs to a structural equation for $Y_i$, as described by \cite{pearl:00,pearl:16}. We further assume that a pre-defined set of ``neighbors'' of $i$, defined as $N(i) \subset \{1, 2, \dots, n\}$, influence $i$. Specifically, their interventions will influence the outcome of $i$. This is represented as $\{Z_j\}_{j \in N(i)} \xdashrightarrow{} Y_i$, where $\xdashrightarrow{}$ signifies either an \emph{indirect} or direct `spillover' causal effect: that is, $Z_j \xdashrightarrow{} Y_i$ is a shorthand notation for possible paths such as $Z_j \rightarrow Y_j \rightarrow Y_i$ or $Z_j \rightarrow Y_i$. We do not explicitly model any connections among outcomes, as our objective function will not require this information.

Finally, we assume that there are no edges from any $Z_i$ into any $X_j$. We can interpret $X$ as features observed prior to intervention, with $Z_i$ possibly changing hidden variables $H_i$ %$X_{ik}^H$ 
not in $X$ but acting as mediators in pathways such as $A_i \rightarrow H_{i} \rightarrow Y_i$ or $Z_i \rightarrow H_{i} \rightarrow Y_i$ ($H_{i}$ may be observed \emph{after} the action takes place, but not conditioned on). The idea explored in the next section is that we choose $\mathbf z$ by first observing all $A$ and $X$ to achieve some measure of fairness within a set of $n$ individuals. Edges from $A_i, X_i$ to $Z$ are omitted for simplicity. Like in the original counterfactual fairness work \cite{kusner:17}, we will assume that there is a model that maps inputs $\{A_i\}_{i = 1}^{n}, \{X_i\}_{i = 1}^{n}, \{Z_i\}_{i = 1}^{n}$ to outputs $\{Y_i\}_{i = 1}^{n}$. For instance, \cite{arbour:16,aronow:17} provide some methods for this task. Our focus is \emph{not} on estimating a causal model: such a model is assumed to be given either by prior experiments or by fitting observational data under causal assumptions. Rather, we focus on defining a measure of fairness for new cases where $\{A_i\}_{i = 1}^{n}, \{X_i\}_{i = 1}^{n}$ have been observed but $\{Y_i\}_{i = 1}^{n}$ has not occurred yet. %Instead of choosing predictors $\hat Y_1, \dots\, \hat Y_n$, we choose actions $Z_1, \dots, Z_n$.

\begin{wrapfigure}{R}{0.5\textwidth}
\begin{center}
\vspace{-3ex}
\centerline{\includegraphics[width=\linewidth]{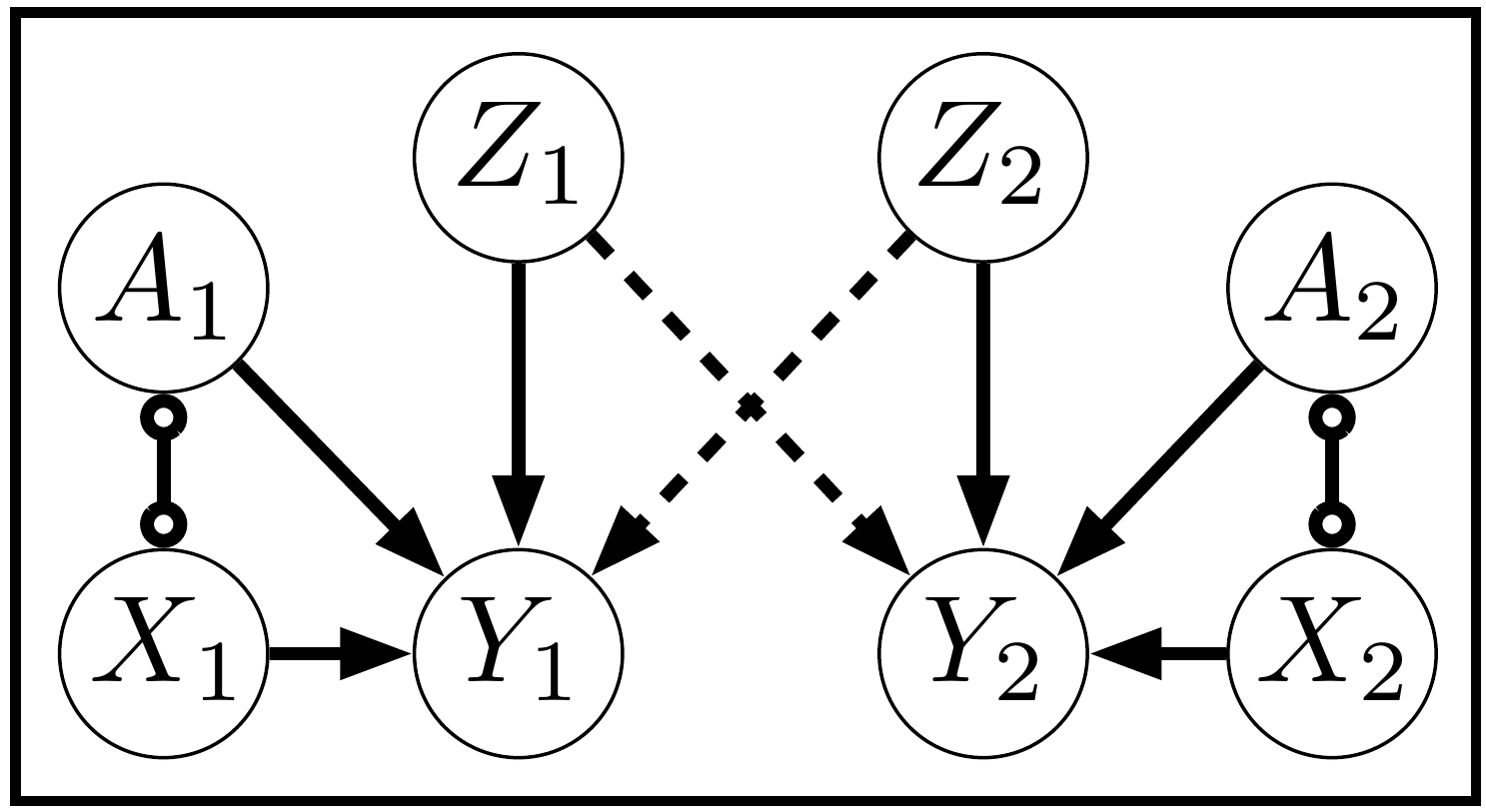}}

  \caption{\small An example causal diagram with action vertices $Z$, outcomes $Y$, features $X$, and sensitive group $A$. Here, individuals $1,2$ are neighbors of each other, thus their interventions may `spillover' in possibly indirect ways, as shown by the dashed arrows. Edges $\circ$-$\circ$ represent arbitrary causal connections between $X_i$ and $A_i$. We omit any edges between $Y_1$ and $Y_2$ for simplicity. \label{fig:example}}
\end{center}
\vspace{-5ex}
\end{wrapfigure}

Figure~\ref{fig:example} shows a simple example where two individuals are neighbors and thus, there is possible interference between their individual interventions.

%\begingroup
%\setlength{\intextsep}{0pt}
%\endgroup

\subsection{$\tau$-Controlled Counterfactual Privilege}

%In an ideal world, there would exist at least one treatment assignment $\mathbf Z = \mathbf z$ such that $Y_i(a, \mathbf z) = Y_i(a', \mathbf z)$ for all $i, a, a'$. For instance, if the causal graph was $A \rightarrow X \rightarrow Y$, and $Z_i$ was a perfect intervention of $X_i$ that breaks the link between $A$ and $X$. In many social sciences problems, such an intervention is a mathematical idealization, never a reality. For instance, if $A$ is ``race,'' $X$ is ``wealth'' and $Y$ is ``person defaults on a loan,'' then an intervention that completely overrides the link $A \rightarrow X$ is impractical, at least in the short-term.

%Even if $Y_i(a, \mathbf z) = Y_i(a', \mathbf z)$ was attainable by some $\mathbf z$,
Without perfect interventions we cannot guarantee $Y_i(a, \mathbf z) = Y_i(a', \mathbf z)$ via some $\mathbf z$. But even if this was possible,
we must still specify how $Y_i(a, \mathbf z)$ is preferable to $Y_i(a, \mathbf 0)$: for instance, it is undesirable to have a policy that removes unfairness by crashing the economy to ensure everyone has zero income. Unlike the prediction problem that just attempts to reconstruct $Y_i(a, \mathbf 0)$, in this control problem we need to consider which outcomes are desirable. Assuming $Y$ is a probabilistic outcome, we proceed by choosing: i) a summary of the distribution of $Y$ that we want to control; ii) an objective function; iii) an appropriate notion of \emph{approximate control of unfairness}. We will use expected values to summarize the distribution of each $Y$, and assume without loss of generality that $Y$ is encoded such that larger values are preferable.

\paragraph{Objective.}
Our goal is to assign (binary) interventions $\mathbf z$ to maximize the sum of expected outcomes over individuals subject to a maximum budget $B$:
\begin{align}
% \mathbf z^\star \equiv \mathrm{argmax}_{\mathbf z} \sum_{i = 1}^n \mathbb E[Y_i(\mathbf z)\ |\ A_i = a_i, X_i = x_i],  %\\
% \;\;\; s.t., \;\sum_{i=1}^n z_i \leq B \label{eq:objective0}
\mathbf z^\star \equiv \mathrm{argmax}_{\mathbf z} &\;\sum_{i = 1}^n \mathbb E[Y_i(\mathbf z)\ |\ A_i = a_i, X_i = x_i], \label{eq:objective0} \\
\;\;\; s.t., &\;\sum_{i=1}^n z_i \leq B  \nonumber
\end{align}
\noindent where $a_i$, $x_i$ are the factual realizations of $A$ and $X$, for individual $i$. As discussed previously, this conditional expectation is assumed to be given by a pre-defined causal model for relational data, well-defined for our target outcomes regardless of the neighborhood of each individual \cite{arbour:16,aronow:17}. 

\paragraph{Constraints: Bounding Group Privileges.}
With a causal graph we can define privilege as having a better outcome \emph{because} of ones value of the protected attribute, i.e. $\mathbb E[Y_i(a, \mathbf 0)] > \mathbb E[Y_i(a', \mathbf 0)]$. Based on this, we might consider interventions $\mathbf z$ inducing ``approximate fairness'' $\mathbb E[Y_i(a, \mathbf z)] \approx \mathbb E[Y_i(a', \mathbf z)]$, for instance by enforcing $|\mathbb E[Y_i(a, \mathbf z)] - \mathbb E[Y_i(a', \mathbf z)]| < \epsilon$ for some predefined $\epsilon$. This does not clarify what happens to features $X_i$, which might lie on the pathway between $A_i$ and $Y_i$ and cannot be simply conditioned on \cite{kusner:17}. Moreover, since the problem is a maximization problem, it is less clear why bounding absolute differences is sensible. Before we present a formal definition, let us introduce a motivating example below.

%\begin{example}[frametitle=Some Headlinetext]
% ,breakable
\begin{tcolorbox}[enhanced jigsaw,pad at break*=1mm,
  colback=blue!5!white,colframe=blue!75!black,title=Example: Housing Subsidies]
\noindent \emph{Consider two individuals, 1 and 2, such that $\{X_1, X_2\}$ are their professional qualifications in some quantitative scale, and $Y_i(a_i, [z_1, z_2])$ are their counterfactuals of interest: total income in the next 10 years, in thousands of dollars. Suppose that $Z_i = 1$ corresponds to the action ``individual $i$ gets a subsidy to move to a neighborhood with convenient transport links'', and suppose our budget for this program is such that $z_1 + z_2 \leq 1$, for $z_i \in \{0, 1\}$. Individual 1 is of a minority group $A_1 = b$ and Individual 2 is of majority group $A_2 = w$.}

\emph{Suppose that there is discrimination in society against group $b$ and that there is no interference between individuals. In particular, assume the structural equation}:
\begin{center}
$Y_i = X_i + 100Z_i + 50Z_i \times \mathbb{I}(A_i = w) + U_i$,
\end{center}
\emph{where $\mathbb I(\cdot)$ is the indicator function and the error term $U_i$ has zero mean. In this case, being in group $w$ gives an extra boost if the corresponding individual is given the chance of moving. If $X_1 < X_2 + 50$, the solution to this optimization problem is to set $Z_2 \!=\! 1$. Even if Individual 1 is more qualified than Individual 2 by up to 50 units, maximizing the ``total well-being'' $\mathbb E[Y_1(z_1)\ |\ b, x_1] + \mathbb E[Y_2(z_2)\ |\ w, x_2]$ still favors the individual in the privileged group.}\\

\emph{The problem is amplified if there is interference, for instance, if giving the treatment to Individual 2 increases the chances of the neighborhood of Individual 1 having fewer people of group $w$, leading to a fall of property prices and neighborhood decay. A model can capture that by adding both $Z_1$ and $Z_2$ to each structural equation, penalizing $Y_i$ for each interaction $Z_i \times \mathbb I(A_i = w)$ such as the following}: 
\begin{center}
$Y_i = X_i + 100Z_i + 50Z_i \times \mathbb{I}(A_i \!=\! w) - 10\mathbb I(Z_i \!=\! 0) \times \sum_j \{Z_j \times \mathbb I(A_j \!=\! w)\} + U_i$. $\Box$
\end{center}
%\end{example}
%  \lipsum[1-12]
\end{tcolorbox}

In the above example, the structural equation for $Y_i$ is assumed to be a fact of society. We cannot change the equation, but we can change its inputs. In particular, we are interested in  \emph{bounded privilege constraints}. If we adopt the constraints:
\begin{equation}
\mathbb E[Y_i(a_i, \mathbf z)\ |\ A_i = a_i,  X_i = x_i] - \mathbb E[Y_i(a', \mathbf z)\ |\ A_i = a_i,  X_i = x_i] < \tau,
\label{eq:privilege}
\end{equation}
for some $\tau \!>\! 0$ and all $a'$ in the domain of $A$, and $i \!\in\! \{1, \ldots, n\}$, we exclude treatment assignments that allow an individual $i$ to gain more than $\tau$ units in expectation due to the interaction of $\mathbf z$ and $A_i$. The interpretation of constraint (\ref{eq:privilege}) is that if individual $i$ has an increase in (expected) outcome that is due (by at least a margin $\tau$) to belonging to group $a_i$, then this is defined as \emph{unfair privilege}. 

In general, the constraint in eq.~\eqref{eq:privilege} requires full-knowledge of the specific form of all structural equations\footnote{Depending on the causal graph, it may be possible to identify the desired functionals without complete knowledge of the structural equations \cite{nabi2017fair}.}. For example, if some $X_{i,[k]}$ is a descendant of $A_i$, then in general $X_{i,[k]}(a_i) \neq X_{i,[k]}(a')$. To avoid that, we can exclude all descendants of $A_i$ (but $Y_i$, which is the outcome to be predicted instead of observed evidence) and fit a model that will not require any structural equation except for the outcome. Note that this assumes we can block any confounding between $A$ and $Y$ (which can be done nonparametrically either by randomized controlled trials or knowledge of the causal graph combined with particular adjustments \cite{pearl:00,sgs:93}). Thus we propose a variation of the above constraint:
%A variation of the above constraint is:
\begin{equation}
\underbrace{
\mathbb E_{\mathcal M^\prec}[Y_i(a_i, \mathbf z)\ | \ A_i = a_i, X_i^{\prec} = x_i^{\prec}] - 
\mathbb E_{\mathcal M^\prec}[Y_i(a', \mathbf z)\ |\ A_i = a_i, X_i^{\prec} =  x_i^{\prec}]}_{G_{ia'}} < \tau,
\label{eq:privilege_simple}
\end{equation}
\noindent where $X_i^{\prec}$ is the subset of $X_i$ that are non-descendants of $A_i$ in the causal graph, and $\mathcal M^\prec$ is a causal model that excludes all observed non-descendants of $A$ but $Y$. 
%The motivation for constraints $G_{ia'}$ as opposed to (\ref{eq:privilege}) is that (in general)\footnote{Depending on the causal graph, it may be possible to identify the desired functionals without complete knowledge of the structural equations \cite{nabi2017fair}.} the former requires full-knowledge of the specific form of all structural equations: for example, if some $X_{ik}$ is a descendant of $A_i$, then in general $X_{ik}(a_i) \neq X_{ik}(a')$. To avoid that, we can exclude all descendants of $A_i$ (but $Y_i$, which is the outcome to be predicted instead of observed evidence) and fit a model $M^\prec$ that will not require any structural equation except for the outcome, assuming we can block any confounding between $A$ and $Y$ (which can be done nonparametrically either by randomized controlled trials or knowledge of the causal graph combined with particular adjustments \cite{pearl:00,sgs:93}). 
Notice that the objective function (\ref{eq:objective0}) can use all information in $X_i$, since there is no need to propagate counterfactual values of $A_i$. Hence, this formulation uses two structural equations for the outcome $Y$, one including $X_i$ and one including $X_i^\prec$. The advantage of (\ref{eq:privilege_simple}) is not requiring structural equations for any variables other than $Y$, which in general would require assumptions that cannot be tested even with randomized controlled trials \cite{loftus:18,kusner:17}. In contrast, the objective function and constraints (\ref{eq:privilege_simple}) can at least in principle be estimated by experiments. The fair optimization problem is therefore:
\begin{align}
\max_{z_1, \ldots, z_n} \sum_{i=1}^n&\; \mathbb E[Y_i(\mathbf z)\ |\ A_i = a_i, X_i = x_i] \label{eq:opt}\\\
s.t., \;\sum_{i=1}^n&\; z_i \leq B \nonumber \\
&\; G_{ia'} \leq \tau \;\;\;\; \forall a' \in \mathcal A, \; i \in \{1,\ldots,n\}, \nonumber
\end{align}
\noindent where $\mathcal A$ is the domain of $A$ and $\tau \!>\! 0$. %This formulation is compatible with other constraints, such as $\sum_{A_i = b} z_i \approx \sum_{A_i = w} z_i$, which can be added depending on the context. 
We call a treatment assignment satisfying the constraints above as \emph{$\tau$-controlled counterfactual privilege}.

\paragraph{The Optimization Framework}
As the difference between an approximate solution and a globally optimal solution could mean the difference between a fair and an unfair solution, and as the problems we are interested in are often only hundreds or thousands of interventions, we propose an optimization procedure to solve it exactly. Our formulation will accommodate any function form for the structural equation for $Y$. To do so, we formulate eq.~\eqref{eq:opt} as a mixed-integer-linear-program ({\sc milp}). To avoid fractional solutions from the {\sc milp} for intervention set $\mathbf{z}$, we use integer constraints to enforce that each intervention $z_i$ is binary in the final solution. Given a set of $n$ individuals, we assume that for each individual $i$ there are at most $K$ other neighbor individuals $N(i)$ whose interventions interfere on their outcome $Y_i$. Let these interventions be called $\mathbf{z}_{N(i)}$. We begin by introducing a fixed auxiliary matrix $\mathbf{E} \in \{0,1\}^{(2^K,K)}$. Each row $\mathbf{e}_j$ corresponds to one of the possible values that $\mathbf{z}_{N(i)}$ can take (i.e, all possible $K$-length binary vectors). 

%We rewrite the objective of eq.~\eqref{eq:objective} to sum over all possible $\mathbf{z}_{N(i)}$
%is a possible interference vector, that indicates which of the $K$ neighbors interferes on $Y_i$, out of $2^k$ possibilities (i.e., this row is a multi-hot vector).
Additionally we introduce a matrix $\mathbf{H} \subseteq \{0,1\}^{(n, 2^K)}$ where each row $\mathbf{h}_i$ indicates for individual $i$, which of the $2^K$ possible neighbor interferences affect $Y_i$ (i.e., each row is a 1-hot vector). We will optimize $\mathbf{H}$ jointly with $\mathbf{z}$. This allows us to rewrite the objective of eq.~\eqref{eq:opt} as: $\sum_{i = 1}^n \sum_{j=1}^{2^k} {h}_{i,j}\mathbb E[Y_i(\mathbf z_{N(i)}= \mathbf{e}_j) \ |\ A_i = a_i, X_i = x_i]$. Note that we introduce a sum over all possible $\mathbf{z}_{N(i)}$ and use $\mathbf{H}$ to indicate which element of this sum is non-zero. We can rewrite the fairness constraints in a similar way.  To ensure that each row $\mathbf{h}_i$ agrees with the actual $\mathbf{z}_{N(i)}$ we enforce the following constraints: $\mathbb{I}[\mathbf{e}_j = 1] h_{i,j} \leq \mathbf{z}_{N(i)}$ and $\mathbb{I}[\mathbf{e}_j = 0] h_{i,j} \leq 1 - \mathbf{z}_{N(i)}$, where $\mathbb{I}$ is the indicator function that operates on each element of a vector.
The first constraint ensures that the non-zero entries of $\mathbf{e}_j$ are consistent with $\mathbf{z}_{N(i)}$ via $h_{i,j}$, and the second ensures the zero entries agree. Finally, to ensure that each row of $\mathbf{H}$ is 1-hot we introduce the constraint $\sum_{j=1}^{2^K} h_{i,j}=1$ for all $i$. This yields the following optimization program:
\begin{align}
&\max_{{\bf H}, {\bf z} \in \{0,1\}} \sum_{i = 1}^n \sum_{j=1}^{2^K} { h}_{i,j}\mathbb E[Y_i(\mathbf z_{N(i)}=\mathbf{e}_j) \ |\ A_i = a_i, X_i = x_i], \label{eq:objective} \\
s.t.& \;\sum_{j=1}^{2^K} {h}_{i,j} \Big(\mathbb{E}[Y_i(\mathbf z_{N(i)}\!=\! \mathbf{e}_j)\ |\ A_i \!=\! a_i, X_i^{\prec} \!=\! x_i^{\prec}]-\mathbb{E}[Y_i(\mathbf z_{N(i)} \!=\! \mathbf{e}_j)\ |\ A_i \!=\! a', X_i^{\prec} \!=\! x_i^{\prec}] \Big) < \tau, \forall a', i \nonumber\\
& \mathbb{I}[\mathbf{e}_j = 1]  h_{i,j} \leq \mathbf{z}_{N(i)}, \,\,\,\,\,\,\,\,\,\,\,\,\forall i,j %\text{ s.t. } \mathbf e_l(j)=1
\nonumber\\
& \mathbb{I}[\mathbf{e}_j = 0] h_{i,j} \leq 1- \mathbf{z}_{N(i)}, \,\, \forall i,j %\text{ s.t. } \mathbf e_l(j)=0
\nonumber\\
% Equivilent to the previous form. 
%& \mathbf h_{i,j} \leq (1-{\bf e}_l(j))+(2{\bf e}_l(j) -1)\mathbf z_{i,l\,} \forall i,j,l \nonumber\\
&\sum_{j=1}^{2^K} h_{i,j}=1,\,\,\,\, \forall i %\nonumber \\ %, 
  \,\,\,\,\,\,\,\,\,\sum_{i=1}^n z_i \leq B. \nonumber %\\
%&\sum_{i=1}^n z_i \leq B. \nonumber
%&\forall \mathbf{H} \in [0,1]. %\text{ and }  \forall z\in \{0,1\}.
\nonumber
\end{align}

\paragraph{Path-specific and Multiple-model Variants.} Following the original exposition of \cite{kusner:17}, we have described the case where all paths from $A$ to $Y$ in the causal graph carry a notion of unfairness. It is possible to extend the constraints above to ``path-specific'' effects, a concept exploited by other causal formalisms such as \cite{kilbertus2017avoiding,kusner:17,nabi2017fair,zhang2018fairness}. Although the extension to path-specific counterfactual fairness is a natural one, its notation can get cumbersome \cite{loftus:18,chiappa:18}, and as such we defer the discussion to a longer version of this paper. Likewise, \cite{russell:17} discuss how to accommodate multiple competing models.

\paragraph{Solution Paths, Feasibility and Ineffective Interventions.} A practitioner does not need to commit herself to a single choice of $\tau$. We may interpret this problem as multiple-objective optimization problem that maximizes with respect to $\mathbf z$ and minimizes with respect to $\tau$, which leads naturally to exploring a whole solution path of possible values of $\tau$ and reporting the corresponding trade-off. In particular, it is possible that $\tau$ must be relatively large for a problem to be feasible. For instance, if the structural equation for $Y$ is additive on $Z_i$, such as $Y_i = \mathbb{I}(A_i = w) + Z_i + U_i$, then no solution will be feasible for $0 \leq \tau < 1$. Moreover, as there is no interaction between $A_i$ and $Z_i$, solutions will be trivial in the sense that they cannot reduce the gap among the counterfactuals compared to $\mathbf z = \mathbf 0$. This is not a problem due to the definition of $\tau$-controlled counterfactual privilege: it is solely the consequence of having an ineffective class of interventions, an issue that cannot be solved by an algorithm but by real-world design.

%\begingroup
%\setlength{\intextsep}{0pt}
\begin{wrapfigure}{R}{0.4\textwidth}
\begin{center}
\vspace{-10ex}
\centerline{\includegraphics[width=\linewidth]{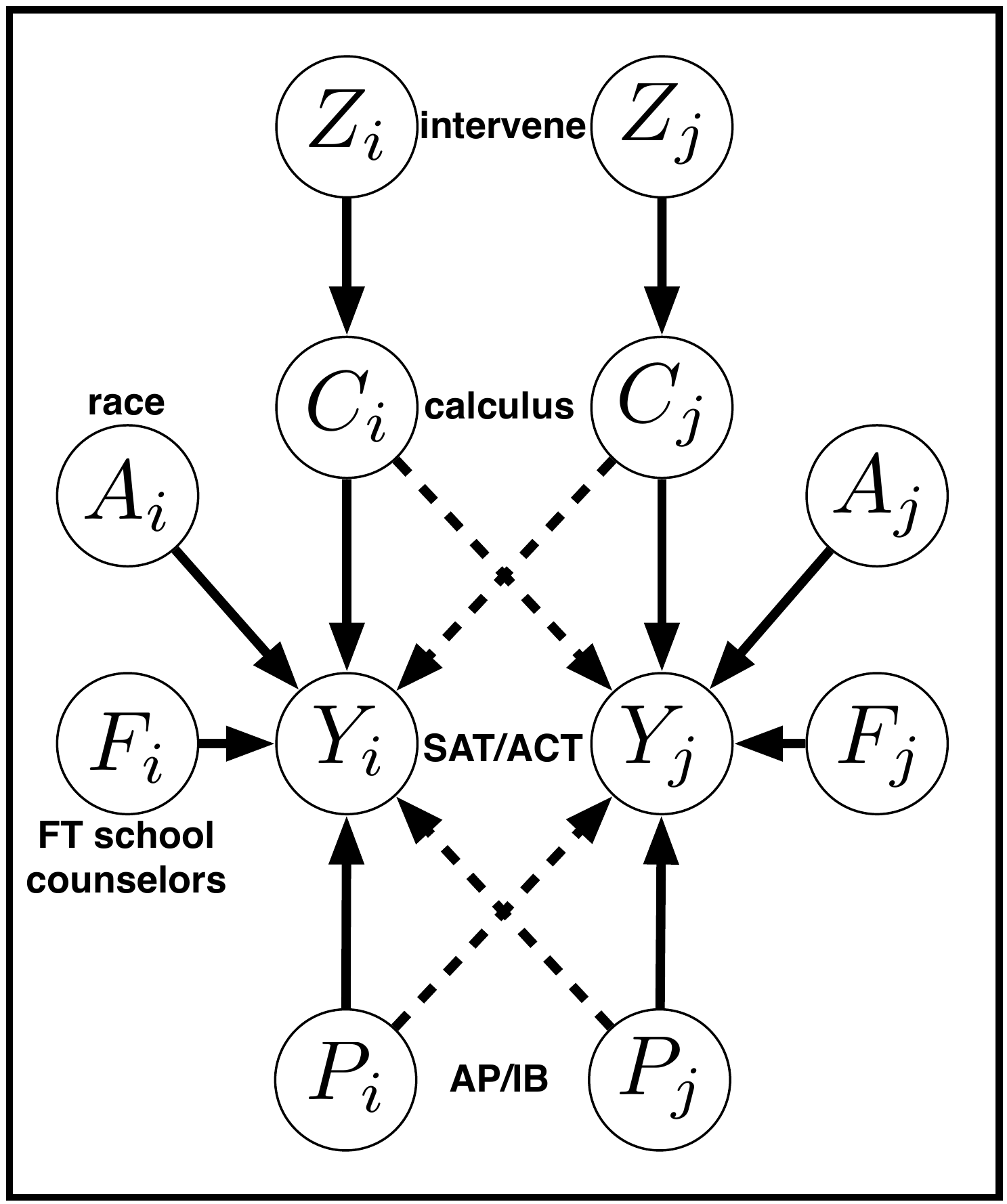}}
  \caption{\small The causal model for the NYC school dataset.   \label{fig:school_model}}
\end{center}
\vspace{-8ex}
\end{wrapfigure}
%\endgroup

%         design a hypothetical intervention 
% TODO: describe dataset:
% - show table with 3 example rows
% - plot 25 example schools on a map of NYC (maybe use R?)
% - show graphical model for this example
% - write functional equations
% OUTLINE:
% - data
% - model
% - results
\section{Experiments}
We now demonstrate our fair allocation of interventions on a real-world dataset. %All code and data to replicate the experiments will be made available\footnote{\url{http://anonymized}}.

\paragraph{Dataset.}
We compiled a dataset on $345$ high schools from the New York City Public School District, largely from the Civil Rights Data Collection (CRDC)\footnote{\url{https://ocrdata.ed.gov/}}. The CRDC collects data on U.S. public primary and secondary schools to ensure that the U.S. Department of Education's financial assistance does not discriminate `on the basis of race, color, national origin, sex, and disability.' This dataset contains 
%This dataset contains information about each school's demographic distributions of race, sex, and disability. Additionally, it has information about the courses offered by the school, the school staff, and student outcomes. %Table~\ref{table.school} shows a snapshot of the dataset and the variables we will use in our experiment including
%\begin{itemize}
demographic information, 
\emph{Full-time School Counselors} ($F$): the number of full-time counselors employed at school (fractional values indicate part-time work),
\emph{AP/IB} ($P$): whether the school offers Advanced Placement (AP) or International Baccalaureate (IB) classes, \emph{Calculus} ($C$): whether the school offers Calculus courses, and 
\emph{SAT/ACT-taking} ($Y$): the percent of students who take the college entrance examinations, the SAT and/or the ACT. For simplicity we assign each school a race $A$ according to its majority race out of the possible groups: \emph{black}, \emph{Hispanic}, \emph{white}.
%\end{itemize}

\paragraph{Setup.} In this experiment, we imagine that the U.S. Department of Education wishes to intervene to offer financial assistance to schools to hire a Calculus teacher, a class that is commonly taken in the U.S. at a college level. The goal is to increase the number of students that are likely to attend college, as measured by the fraction of students taking the entrance examinations (via \emph{SAT/ACT-taking}). It is reasonable to assume that this intervention is exact. Specifically, if the intervention is given to school $i$, i.e., $Z_i\!=\!1$, then we assume that the school offers Calculus, i.e., $C_i\!=\!1$. Without fairness considerations, the Department would simply assign interventions to maximize the total expected percent of students taking the SAT/ACT until they reach their allocation budget $B$. However, to ensure we allocate interventions to schools that will benefit \emph{independent of their societal privilege due to race} we will learn a model using the fairness constraints described in eq.~\eqref{eq:opt}. We begin by formulating a causal model that describes the relationships between the variables.

\begin{figure*}[t]
\centering
\includegraphics[width=1.0\textwidth]{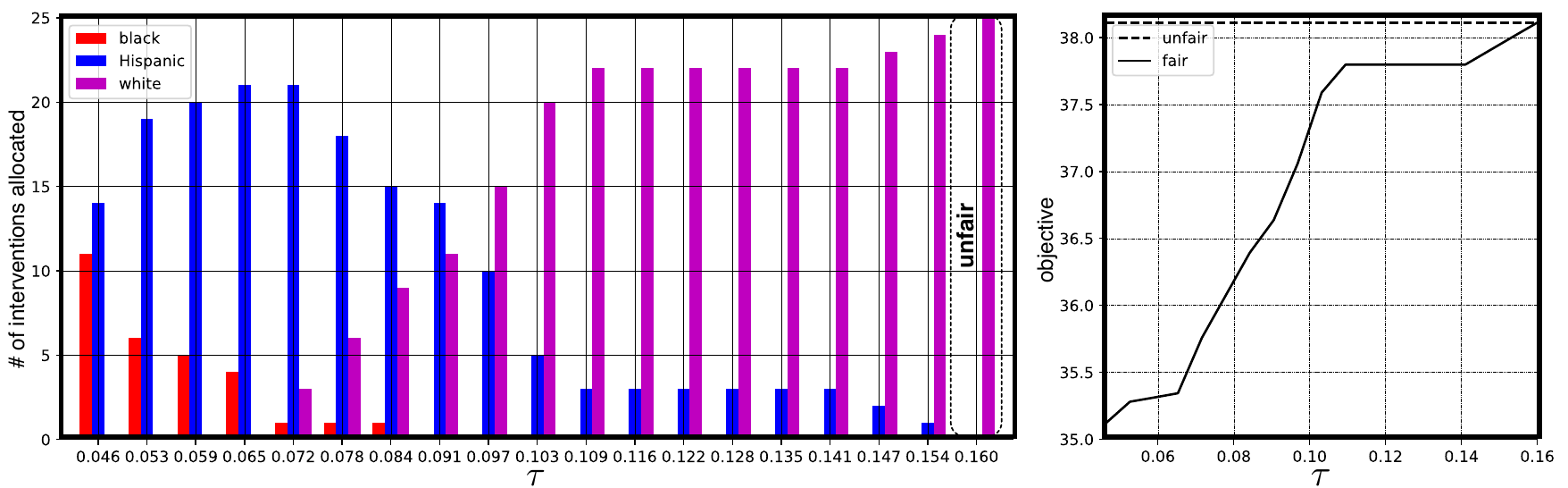}
\vspace{-3ex}
%\caption{Results of the}
\caption{Results for allocating fair interventions for the NYC school dataset. See text for details.}
\label{fig:school_plot_both}
\end{figure*}

\paragraph{Causal model.}
The structure of the causal model we propose is shown in Figure~\ref{fig:school_model} (a subset of the graph is shown for schools $i$ and $j$). Recall that technically $Z_i$ does not directly effect observable variables. $C_i$ is hidden to the extent that its value is only observable after the action takes place. All variables directly affect the outcome $Y_i$ (SAT/ACT-taking). Frequently schools will allow students from nearby schools to take classes that are not offered at their own school. Thus we model both the Calculus class variables $C$ and the AP/IB class variables $P$ as affecting the outcome of students at neighboring schools. Specifically, we propose the following structural equations for $Y$ with interference:
%. First a model with \emph{linear interference}:% interact to affect the outcome of different schools.
% \begin{align}
% %j \;|\; j \in N(i), Z_j = 1
% \mathbb{E}[Y_i(\mathbf{z}) \mid a_i, p_i, f_i] = \alpha_{[a_i]} \!\!\!\sum_{\substack{j \in N(i)\\ s.t., z_j = 1}} s(i,j) C_j(z_j) + \beta_{[a_i]} \!\!\sum_{j \in N(i)} s(i,j) p_j + \gamma_{[a_i]} f_i + \theta_{[a_i]} \label{model:linear}
% \end{align}
%and \emph{max interference}:
\begin{align}
%j \;|\; j \in N(i), Z_j = 1
\mathbb{E}[Y_i(\mathbf{z}) \mid a_i, p_i, f_i] = \alpha_{[a_i]} \!\!\!\max_{\substack{j \in N(i)\\ s.t., z_j = 1}} s(i,j) C_j(z_j) + \beta_{[a_i]} \max_{j \in N(i)} s(i,j) p_j + \gamma_{[a_i]} f_i + \theta_{[a_i]} \label{model:max}
\end{align}
where $C_j(z_j) = z_j$, $N(i)$ refers to the nearby schools of school $i$ (including $i$), and $s(i,j)$ is the similarity of schools $i$ and $j$. We construct both $N(i)$ and $s(i,j)$ using GIS coordinates for each school in our dataset\footnote{\url{https://data.cityofnewyork.us/Education/School-Point-Locations/jfju-ynrr}}: $N(i)$ is the nearest $5$ schools to school $i$ and $s(i,j)$ is the inverse distance in GIS coordinate space. We fit the parameters $\boldsymbol{\alpha}, \boldsymbol{\beta}, \boldsymbol{\gamma}, \boldsymbol{\theta}$ via maximum likelihood, assuming a Gaussian noise model for $Y$.

\begin{figure*}[t]
\centering
\includegraphics[width=1.0\textwidth]{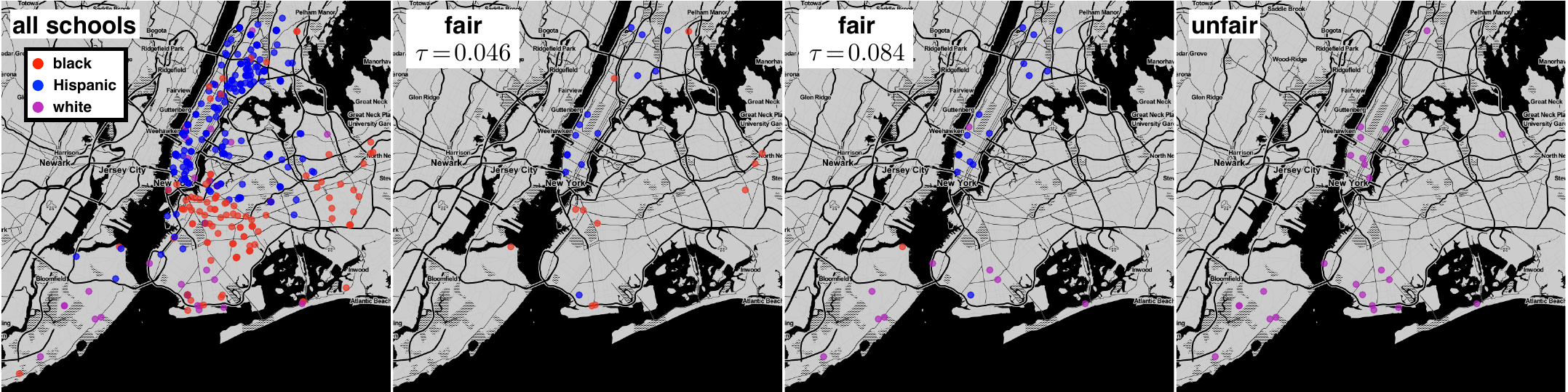}
\vspace{-3ex}
%\caption{Results of the}
\caption{The left-most plot shows the locations of the $345$ New York City High Schools, and their majority race. The remaining plots show the allocations of interventions for each policy.}
\label{fig:maps}
\end{figure*}

\paragraph{Results.}
To evaluate the effect on SAT/ACT-taking when intervening on Calculus courses we start with null allocation vector $\mathbf{z}\!=\!0$ (i.e., no school has a Calculus course). We then solve the optimization problem in eq.~\eqref{eq:opt} with the structural equation for $Y$ in eq.~\eqref{model:max}, and a budget $B$ of $25$ schools. The results of the fair model is shown in Figure~\ref{fig:school_plot_both}. The left plot shows the number of interventions allocated to schools by race. The right plot shows the objective value achieved by the fair and unfair (unconstrained) models. On the far right of the left plot is the unfair allocation. In this case, all interventions are given to predominantly white schools. %As $\tau$ is reduced, more and more interventions are allocated to Hispanic and black schools. %This is because for the linear model the weights $\boldsymbol{\alpha}$ strongly influence the contribution of an intervention. While the weight $\alpha_{\textrm{white}}$ is largest ($0.154$), the next closest is $\alpha_{\textrm{Hispanic}}$ ($0.047$), thus it is allocated when the fairness constraints are active. %Notice that the value of the fair objective in the right plot roughly tracks the number of white schools given interventions.
%Figure~\ref{fig:school_plot_both} (bottom) shows the results of the max model eq.~\eqref{model:max}. 
When $\tau$ is small both predominantly black and Hispanic schools receive allocations because these schools benefit the least from their race. As $\tau$ is increased Hispanic school allocations increase, then decrease as white schools are allocated.  %Different from the linear model the fair objective is not as directly tied to the number of white schools particularly for $\tau \in [0.046, 0.065]$.

Figure~\ref{fig:maps} shows how each policy allocates interventions on a map of New York City. The fair policy ($\tau \!=\! 0.046$, the first set of bars in Figure~\ref{fig:school_plot_both}) assigns interventions to predominantly Hispanic and black schools that have high utility because of things \emph{not due to race}. Some of these schools are in neighborhoods with low median income such as North Bronx and North Manhattan\footnote{\url{http://uk.businessinsider.com/new-york-city-income-maps-2014-12?r=US&IR=T}}. As $\tau$ is increased ($\tau \!=\! 0.084$, the seventh set of bars in Figure~\ref{fig:school_plot_both}) the allocation includes more majority white schools, less black schools, and roughly the same number of Hispanic schools. There are more allocations in Brooklyn and Queens. The only intervention to a majority black school kept is to the one in the racially-diverse St. George neighborhood of Staten Island\footnote{\url{https://www.nytimes.com/interactive/2015/07/08/us/census-race-map.html}}. The unfair policy assigns interventions to schools in traditionally white neighborhoods including lower Manhattan, and lower Brooklyn, and all allocations are to white schools. Curiously, many of the interventions made by the unfair model in Manhattan are nearby those made by the fair models to majority Hispanic schools.

\section{Conclusion}
%Much of the past work in algorithmic fairness does not consider how to change the underlying system that gives rise to unfair predictions. This work addresses underlying unfairness in the world by learning how to allocate interventions that alter the underlying causal graph. While often it is not possible to intervene directly on the outcome of interest we learn interventions that largely mitigate the preference for certain groups to have good outcomes, thus reducing privileges ingrained in society.
In this paper we depart from much of the past work on algorithmic fairness by focusing on designing fair interventions to change the underlying system rather than on fair prediction. 
We make use of structural causal models to encode the effects of public policy interventions, including potential interference. 
For concreteness we pursued a particular optimization problem where the intervention has a budget constraint, but the approach can be used for other kinds of optimization problems. 
%requires careful causal modeling. Once done, we make use of causal tools for interventions
%As with other causal approaches to fairness this may be the greatest limitation of our method, but with interventions it is unavoidable, and the causal inference literature provides tools for inferring such models. 
We devised a fairness criterion that allows us to find beneficial interventions while bounding the benefit to individuals caused by being a member of a privileged group, as determined by the causal model. 
We test our method on allocating college-preparatory classes to NYC schools fairly. We found that when the fairness criterion holds, interventions are given to a more racially diverse set of schools.

\bibliography{bibliography}
\bibliographystyle{plainnat}

\end{document}


\maketitle

\begin{table}
\centering
\caption{A snapshot of the NYC high school dataset.}
\label{table.school}
\begin{scriptsize}
\begin{tabular}{cccccccccc}
\toprule
Total & Percent & Percent & Percent & Percent & FT School & Offers AP & Offers & Percent  \\
Students & Black & Hispanic & White & Other & Counselors & and/or IB? & Calculus? & SAT/ACT  \\
  \midrule
$115$ & $84.3\%$ & $8.7\%$ & $3.5\%$ & $3.5\%$ & $1$ & No & No & $3.5\%$ \\ %44
$427$ & $28.3\%$ & $69.1\%$ & $1.6\%$ & $1.0\%$ & $3$ & No & No & $11.5\%$ \\ %18
$369$ & $3.5\%$  & $14.1\%$ & $56.4\%$& $26.0\%$& $1.5$&Yes & Yes & $53.7\%$ \\   %5
\bottomrule
\end{tabular}
\end{scriptsize}
\end{table}

\bibliography{bibliography}
\bibliographystyle{plainnat}